# A Logic for Reasoning about Upper Probabilities

**Joseph Y. Halpern**
Department of Computer Science
Cornell University
Ithaca, NY 14853
halpern@cs.cornell.edu
http://www.cs.cornell.edu/home/halpern

**Riccardo Pucella**
Department of Computer Science
Cornell University
Ithaca, NY 14853
riccardo@cs.cornell.edu

## Abstract

We present a propositional logic to reason about the uncertainty of events, where the uncertainty is modeled by a set of probability measures assigning an interval of probability to each event. We give a sound and complete axiomatization for the logic, and show that the satisfiability problem is NP-complete, no harder than satisfiability for propositional logic.

## 1 INTRODUCTION

Various measures exist that attempt to quantify uncertainty. For many trained in the use of probability theory, probability measures are an obvious choice. However, probability measures have difficulties dealing with certain situations of interest. For instance, while probabilities can model the direct uncertainty of an event happening, it is not always clear how to model higher-order uncertainty, that is, the uncertainty related to the probabilities themselves. Consider a simple example: suppose we have a bag of 100 marbles; we know 30 are red and we know the remaining 70 are either blue or yellow, although we do not know the exact proportion of blue and yellow. If we are modeling the situation where we pick a ball from the bag at random, we need to assign a probability to three different events: picking up a red ball (*red-event*), picking up a blue ball (*blue-event*), and picking up a yellow ball (*yellow-event*). We can clearly assign a probability of .3 to *red-event*, but there is no clear probability to assign to *blue-event* or *yellow-event*.

One way to approach this problem is to represent the uncertainty using a set of probability measures, with a probability measure for each possible proportion of blue and yellow balls. For instance, we could use the set of probabilities $\mathcal{P} = \{\mu_\alpha : \alpha \in [0, .7]\}$, where $\mu_\alpha$ gives *red-event* probability .3, *blue-event* probability $\alpha$, and *yellow-event* probability $.7 - \alpha$. To any set of probabilities $\mathcal{P}$ we can assign a pair of functions, the upper and lower probability measure, that for an event $X$ give the supremum (respectively, the infimum) of the probability of $X$ according to the probability measures in $\mathcal{P}$. These measures can be used to deal with uncertainty in the manner described above, where the lower and upper probability of an event defines a range of probability for that event.[1] Note that this is not the only way to model the situation. An alternative approach, using inner measures, is studied in [Fagin and Halpern 1991].

Given a measure of uncertainty, one can define a logic for reasoning about it. Fagin, Halpern and Megiddo [1990] introduce a logic for reasoning about probabilities, with a possible-worlds semantics that assigns a probability to each possible world. They provide an axiomatization for the logic, which they prove sound and complete with respect to the semantics. They also show that the satisfiability problem for the logic, somewhat surprisingly, is NP-complete, and hence no harder than the satisfiability problem for propositional logic. They moreover show how their logic can be extended to other notions of uncertainty, such as inner measures [Fagin and Halpern 1991] and Dempster-Shafer belief functions [Shafer 1976].

In this paper, we describe a logic for reasoning about upper probability measures, along the lines of the logic introduced in [Fagin, Halpern, and Megiddo 1990]. The main challenge is to derive a provably complete axiomatization of the logic; to do this, we need a characterization of upper probability measures in terms of properties that can be expressed in the logic. Many semantic characterizations of upper probability measures have been proposed in the literature. The characterization of Anger and Lembcke [1985] turns out to be best suited for our purposes. Even though we are reasoning about potentially infinite sets of probability measures, the satisfiability problem for our logic remains NP-complete. Intuitively, we need guess only a small number of probability measures to satisfy any given formula, polynomially many in the size of the formula. Moreover, these probability measures can be taken to be defined on a finite state space, again polynomial in the size of the formula. Thus, we need to basically determine polynomially many values—a value for each probability measure at each state—to decide the satisfiability of a formula.

---

[1] An alternate view of upper probabilities originates from assigning subjective probabilities to events by testing and finding at what odds a person is prepared to bet against them. This gives rise, given suitable assumptions, to an equivalent formulation of lower and upper probability measures [Smith 1961; Walley 1991].

The rest of this paper is structured as follows. In Section 2, we review the required material from probability theory and the theory of upper probabilities. In Section 3, we present the logic and an axiomatization. In Section 4, we prove that the axiomatization is sound and complete with respect to the natural semantic models expressed in terms of upper probability spaces. Finally, in Section 5, we prove that the decision problem for the logic is NP-complete. We leave out the proofs of the more technical results. These can be found in the full version of the paper [Halpern and Pucella 2001].

## 2 CHARACTERIZING UPPER PROBABILITY MEASURES

We start with a brief review of the relevant definitions. Recall that a probability measure is a function $\mu : \Sigma \to [0, 1]$ for $\Sigma$ an algebra of subsets of $\Omega$ (that is $\Sigma$ is closed under complements and unions), satisfying $\mu(\emptyset) = 0$, $\mu(\Omega) = 1$ and $\mu(A \cup B) = \mu(A) + \mu(B)$ for all disjoint sets $A, B$ in $\Sigma$.[2] A probability space is a tuple $(\Omega, \Sigma, \mu)$, where $\Omega$ is a set, $\Sigma$ is an algebra of subsets of $\Omega$ (the measurable sets), and $\mu$ is a probability measure defined on $\Sigma$. Given a set $\mathcal{P}$ of probability measures, let $\mathcal{P}^*$ be the upper probability measure[3] defined by $\mathcal{P}^*(X) = \sup\{\mu(X) : \mu \in \mathcal{P}\}$ for $X \in \Sigma$. Similarly, $\mathcal{P}_*(X) = \inf\{\mu(X) : \mu \in \mathcal{P}\}$ is the lower probability of $X \in \Sigma$. A straightforward derivation shows that the relationship $\mathcal{P}_*(X) = 1 - \mathcal{P}^*(\overline{X})$ holds between upper and lower probabilities, where $\overline{X}$ is the complement of $X$ in $\Omega$. Because of this duality, we restrict the discussion to upper probability measures in this paper, with the understanding that results for lower probabilities can be similarly derived. Finally, an *upper probability space* is a tuple $(\Omega, \Sigma, \mathcal{P})$ where $\mathcal{P}$ is a set of probability measures on $\Sigma$.

---

[2]If $\Omega$ is infinite, we could also require that $\Sigma$ be a $\sigma$-algebra (i.e., closed under countable unions) and that $\mu$ be countably additive. Requiring countable additivity would not affect our results, since we show that we can take $\Omega$ to be finite. For ease of exposition, we have not required it.

[3]In the literature, the term upper probability is sometimes used in a more restricted sense than here. For example, Dempster [1967] uses the term to denote a class of measures which were later characterized as Dempster-Shafer belief functions [Shafer 1976]; belief functions are in fact upper probability measures in our sense, but the converse is not true [Kyburg 1987]. In the measure theory literature, what we call upper probability measures are a special case of *upper envelopes* of measures, which are defined as the sup of sets of general measures, not just probability measures.

We would like a set of properties that completely characterizes upper probability measures. In other words, we would like a set of properties that allow us to determine if a function $f : \Sigma \to [0, 1]$ (for an algebra $\Sigma$ of subsets of $\Omega$) is an upper probability measure, that is, whether there exists a set $\mathcal{P}$ of probability measures such that for all $X \in \Sigma$, $\mathcal{P}^*(X) = f(X)$.

One approach to the characterization of upper probability measures is to adapt the characterization of Dempster-Shafer belief functions; these functions are known to be the lower envelope of the probability measures that dominate them, and thus form a subclass of the class of lower probability measures. By the duality noted earlier, a characterization of lower probability measures would yield a characterization of upper probability measures. The characterization of belief functions is derived from a generalization of the following inclusion-exclusion principle for probabilities (by replacing the equality with an inequality):

$$\mu(\bigcup_{i=1}^{n} A_i) = \sum_{i=1}^{n}(-1)^{i-1}(\sum_{\substack{J \subseteq \{1,\ldots,n\} \\ |J|=i}} \bigcap_{j \in J} A_j).$$

It seems reasonable that a characterization of lower (or upper) probability measures could be derived along similar lines. As we now show, most properties derivable from the inclusion-exclusion principle (which include most of the properties reported in the literature) are insufficient to characterize upper probability measures.

Consider the following "inclusion-exclusion"–style properties (mainly taken from [Walley 1991]). To simplify the statement of these properties, let $\mathcal{P}^{-1} = \mathcal{P}^*$ and $\mathcal{P}^{+1} = \mathcal{P}_*$.

(1) $\mathcal{P}^*(A_1 \cup \cdots \cup A_n) \leq$
    $\sum_{i=1}^{n} \sum_{\{I|=i\}} (-1)^{i+1} \mathcal{P}^{(-1)^i}(\bigcap_{j \in I} A_j)$,

(2) $\mathcal{P}_*(A_1 \cup \cdots \cup A_n) \geq$
    $\sum_{i=1}^{n} \sum_{|I|=i} (-1)^{i+1} \mathcal{P}^{(-1)^{i+1}}(\bigcap_{j \in I} A_j)$,

(3) $\mathcal{P}_*(A \cup B) + \mathcal{P}_*(A \cap B) \leq \mathcal{P}_*(A) + \mathcal{P}^*(B) \leq$
    $\mathcal{P}^*(A \cup B) + \mathcal{P}^*(A \cap B)$,

(4) $\mathcal{P}_*(A) + \mathcal{P}_*(B) \leq \mathcal{P}_*(A \cup B) + \mathcal{P}^*(A \cap B) \leq$
    $\mathcal{P}^*(A) + \mathcal{P}^*(B)$,

(5) $\mathcal{P}_*(A) + \mathcal{P}_*(B) \leq \mathcal{P}_*(A \cap B) + \mathcal{P}^*(A \cup B) \leq$
    $\mathcal{P}^*(A) + \mathcal{P}^*(B)$.

It is easily verified that the above properties hold for upper probability measures. The issue is whether they completely characterize the class of upper probability measures. We show the inherent incompleteness of these properties by proving that they are all derivable from the following simple property, which is by itself insufficient to characterize upper probability measures.

(6) If $A \cap B = \emptyset$, then $\mathcal{P}^*(A) + \mathcal{P}_*(B) \leq \mathcal{P}^*(A \cup B) \leq \mathcal{P}^*(A) + \mathcal{P}^*(B)$.

**Proposition 2.1:** *Property (6) implies properties (1)-(5).*

The following example shows the inusfficiency of Propoerty (6). Let $\mathcal{P}$ be the set of probability measures $\{\mu_1, \mu_2, \mu_3, \mu_4\}$ over $\Omega = \{a, b, c, d\}$ (with $\Sigma$ containing all subsets of $\Omega$) defined on singletons by:

$\mu_1(a) = \frac{1}{4}$　$\mu_1(b) = \frac{1}{4}$　$\mu_1(c) = \frac{1}{4}$　$\mu_1(d) = \frac{1}{4}$

$\mu_2(a) = 0$　$\mu_2(b) = \frac{1}{8}$　$\mu_2(c) = \frac{3}{8}$　$\mu_2(d) = \frac{1}{2}$

$\mu_3(a) = \frac{1}{8}$　$\mu_3(b) = \frac{3}{8}$　$\mu_3(c) = 0$　$\mu_3(d) = \frac{1}{2}$

$\mu_4(a) = \frac{3}{8}$　$\mu_4(b) = 0$　$\mu_4(c) = \frac{1}{8}$　$\mu_4(d) = \frac{1}{2}$

and extended by additivity to all of $\Sigma$. This defines an upper probability measure $\mathcal{P}^*$ over $\Sigma$. Consider the function $v_\epsilon : \Sigma \to [0, 1]$ defined by:

$$v_\epsilon(X) = \begin{cases} \mathcal{P}^*(X) + \epsilon & \text{if } X = \{a, b, c\} \\ \mathcal{P}^*(X) & \text{otherwise} \end{cases}$$

We claim that the function $v_\epsilon$, for small enough $\epsilon > 0$, satisfies property (6), but cannot be an upper probability measure.

**Proposition 2.2:** *For $0 < \epsilon < \frac{1}{8}$, the function $v_\epsilon$ satisfies property (6), but is not an upper probability measure. That is, we cannot find a set $\mathcal{P}'$ of probability measures such that $v_\epsilon = (\mathcal{P}')^*$.*

This example clearly illustrates the need to go beyond the inclusion-exclusion principle to find properties that characterize upper probability measures. As it turns out, various complete characterizations have been described in the literature [Lorentz 1952; Huber 1976; Huber 1981; Williams 1976; Wolf 1977; Giles 1982; Anger and Lembcke 1985]. While all are equivalent in spirit, we focus on the characterization given by Anger and Lembcke [1985], because it is particularly well-suited to the logic presented in the next section. The characterization is based on the notion of *set cover*: a set $A$ is said to be covered $n$ times by a multiset $\{\{A_1, \ldots, A_m\}\}$ of sets if every element of $A$ appears in at least $n$ sets from $A_1, \ldots, A_m$: for all $x \in A$, there exists $i_1, \ldots, i_n$ in $\{1, \ldots, m\}$ such that for all $j \leq n, x \in A_{i_j}$. It is important to note here that $\{\{A_1, \ldots, A_m\}\}$ is a multiset, not a set; the $A_i$'s are not necessarily distinct. (We use the $\{\{\ \}\}$ notation to denote multisets.) An $(n, k)$-*cover of* $(A, \Omega)$ is a multiset $\{\{A_1, \ldots, A_m\}\}$ that covers $\Omega$ $k$ times and covers $A$ $n + k$ times.

The notion of $(n, k)$-cover is the key concept in Anger and Lembcke's characterization of upper probability measures.

**Theorem 2.3:** [Anger and Lembcke 1985] *Let $\Omega$ be a set, $\Sigma$ an algebra of subsets of $\Omega$, and $v$ a function $v : \Sigma \to [0, 1]$. There exists a set $\mathcal{P}$ of probability measures with $v = \mathcal{P}^*$ if and only if $v$ satisfies the following three properties:*

UP1.　$v(\emptyset) = 0$,

UP2.　$v(\Omega) = 1$,

UP3.　*for all integers $m, n, k$ and all subsets $A_1, \ldots, A_m$ in $\Sigma$, if $\{\{A_1, \ldots, A_m\}\}$ is an $(n, k)$-cover of $(A, \Omega)$, then $k + nv(A) \leq \sum_{i=1}^m v(A_i)$.*

We need to strengthen Theorem 2.3 in order to prove the main result of this paper, namely, the completeness of the axiomatization of the logic we introduce in the next section. We show that if the cardinality of the state space $\Omega$ is finite, then we need only finitely many instances of property **UP3**. Notice that we cannot derive this from Theorem 2.3 alone: even if $|\Omega|$ is finite, **UP3** does not provide any bound on $m$, the number of sets to consider in an $(n, k)$ cover of a set $A$. Indeed, there does not seem to be any *a priori* reason why the value of $m$, $n$, and $k$ can be bounded. Bounding this value of $m$ (and hence of $n$ and $k$, since they are no larger than $m$) is the one of the key technical results of this paper, and the necessary foundation of our work.

**Theorem 2.4:** *There exists constants $B_0, B_1, \ldots$ such that if $\Sigma$ is an algebra of subsets of $\Omega$ and $v$ is a function $v : \Sigma \to [0, 1]$, then there exists a set $\mathcal{P}$ of probability measures such that $v = \mathcal{P}^*$ if and only if $v$ satisfies the following properties:*

UPF1.　$v(\emptyset) = 0$,

UPF2.　$v(\Omega) = 1$,

UPF3.　*for all integers $m, n, k \leq B_{|\Omega|}$ and all sets $A_1, \ldots, A_m$, if $\{\{A_1, \ldots, A_m\}\}$ is an $(n, k)$-cover of $(A, \Omega)$, then $k + nv(A) \leq \sum_{i=1}^m v(A_i)$.*

Property **UPF3** is significantly weaker than **UP3**. In principle, checking that **UP3** holds for a given function requires checking that it holds for arbitrarily large collections of sets, even if the underlying set $\Omega$ is finite. On the other hand, **UPF3** guarantees that it is in fact sufficient to look at collections of size at most $B_{|\Omega|}$. This observation is key to the completeness result.

It is not important for our purposes (namely to get completeness of the axiomatization of the logic introduced in the next section) what the actual values of $B_0, B_1, \ldots$ are; it is sufficient for them to exist and be finite. The proof of Theorem 2.4 relies on a Ramsey-theoretic argument that does not provide a bound on the $B_i$s.

## 3   THE LOGIC

The syntax for the logic is straightforward, and is taken from [Fagin, Halpern, and Megiddo 1990]. We fix a set $\Phi_0 = \{p_1, p_2, \ldots\}$ of *primitive propositions*. The set $\Phi$ of *propositional formulas* is the closure of $\Phi_0$ under $\wedge$ and $\neg$. We assume a special propositional formula *true*, and abbreviate $\neg true$ as *false*. We use $p$ to represent primitive propositions, and $\varphi$ and $\psi$ to represent propositional formulas. A *term* is an expression of the form $\theta_1 l(\varphi_1) + \cdots + \theta_k l(\varphi_k)$, where $\theta_1, \ldots, \theta_k$ are reals and $k \geq 1$. A *basic likelihood formula* is a statement of the form $t \geq \alpha$, where $t$ is a term and $\alpha$ is a real. A *likelihood formula* is a boolean combination of basic likelihood formulas. We use $f$ and $g$ to represent likelihood formulas. We use obvious abbreviations where needed, such as $l(\varphi) - l(\psi) \geq a$ for $l(\varphi) + (-1)l(\psi) \geq a$, $l(\varphi) \geq l(\psi)$ for $l(\varphi) - l(\psi) \geq 0$, $l(\varphi) \leq a$ for $-l(\varphi) \geq -a$, $l(\varphi) < a$ for $\neg(l(\varphi) \geq a)$ and $l(\varphi) = a$ for $(l(\varphi) \geq a) \wedge (l(\varphi) \leq a)$. Define the length $|f|$ of the likelihood formula $f$ to be the number of symbols required to write $f$, where each coefficient is counted as one symbol.

We assign a semantics to likelihood formulas through an upper probability space, as defined in Section 2. Formally, an *upper probability structure* is a tuple $M = (\Omega, \Sigma, \mathcal{P}, \pi)$ where $(\Omega, \Sigma, \mathcal{P})$ is an upper probability space and $\pi$ associates with each state (or world) in $\Omega$ a truth assignment on the primitive propositions in $\Phi_0$. Thus, $\pi(s)(p) \in \{\mathbf{true}, \mathbf{false}\}$ for $s \in \Omega$ and $p \in \Phi_0$. Let $[\![p]\!]_M = \{s \in \Omega \;:\; \pi(s)(p) = \mathbf{true}\}$. We call $M$ *measurable* if for each $p \in \Phi_0$, $[\![p]\!]_M$ is measurable. If $M$ is measurable then $[\![\varphi]\!]_M$ is measurable for all propositional formulas $\varphi$. In this paper, we restrict our attention to measurable upper probability structures. Extend $\pi(s)$ to a truth assignment on all propositional formulas in a standard way, and associate with each propositional formula the set $[\![\varphi]\!]_M = \{s \in \Omega \;:\; \pi(s)(\varphi) = \mathbf{true}\}$. An easy structural induction shows that $[\![\varphi]\!]_M$ is a measurable set. If $M = (\Omega, \Sigma, \mathcal{P}, \pi)$, let

$$M \models \theta_1 l(\varphi_1) + \cdots + \theta_k l(\varphi_k) \geq \alpha \text{ iff}$$
$$\theta_1 \mathcal{P}^*([\![\varphi_1]\!]_M) + \cdots + \theta_k \mathcal{P}^*([\![\varphi_k]\!]_M) \geq \alpha$$
$$M \models \neg f \text{ iff } M \not\models f$$
$$M \models f \wedge g \text{ iff } M \models f \text{ and } M \models g.$$

Note that the logic can express lower probabilities: it follows from the duality between upper and lower probabilities that $M \models -l(\neg\varphi) \geq \beta - 1$ iff $\mathcal{P}_*([\![\neg\varphi]\!]_M) \geq \beta$.[4]

Consider the following axiomatization $\mathbf{AX}^{up}$ for likelihood formulas, which we prove sound and complete in the next

---

[4]Another approach, more in keeping with [Fagin, Halpern, and Megiddo 1990], would be to interpret $l$ as a lower probability measure. On the other hand, interpreting $l$ as an upper probability measure is more in keeping with the literature on upper probabilities.

---

section. As in [Fagin, Halpern, and Megiddo 1990], $\mathbf{AX}^{up}$ is divided into three parts, dealing respectively with propositional reasoning, reasoning about linear inequalities, and reasoning about upper probabilities.

*Propositional reasoning*

**Taut**. All instances of propositional tautologies,

**MP**. From $f$ and $f \implies g$ infer $g$.

*Reasoning about linear inequalities*

**Ineq**. All instances of valid formulas about linear inequalities (see below).

*Reasoning about upper probabilities*

**L1**. $l(\mathit{false}) = 0$,

**L2**. $l(\mathit{true}) = 1$,

**L3**. $l(\varphi) \geq 0$,

**L4**. $l(\varphi_1) + \cdots + l(\varphi_m) - nl(\varphi) \geq k$ if
$\varphi \Rightarrow \bigvee_{J \subseteq \{1,\ldots,m\},\, |J|=k+n} \bigwedge_{j \in J} \varphi_j$ and
$\bigvee_{J \subseteq \{1,\ldots,m\},\, |J|=k} \bigwedge_{j \in J} \varphi_j$ are propositional tautologies.

**L5**. $l(\varphi) = l(\psi)$ if $\varphi \Leftrightarrow \psi$ is a propositional tautology.

The only difference between $\mathbf{AX}^{up}$ and the axiomatization for reasoning about probability given in [Fagin, Halpern, and Megiddo 1990] is that the axiom $l(\varphi \wedge \psi) + l(\varphi \wedge \neg\psi) = l(\varphi)$ in [Fagin, Halpern, and Megiddo 1990], which expresses the additivity of probability, is replaced by **L4**. Although it may not be immediately obvious, **L4** is the logical analogue of **UP3**. To see this, first note that $\{\{A_1, \ldots, A_m\}\}$ covers $A$ $m$ times if and only if $A \subseteq \bigcup_{J \subseteq \{1,\ldots,m\},\, |J|=n} \bigcap_{j \in J} A_j$. Thus, the formula $\varphi \Rightarrow \bigvee_{J \subseteq \{1,\ldots,m\},\, |J|=k+n} \bigwedge_{j \in J} \varphi_j$ says that $\varphi$ (more precisely, the set of worlds where $\varphi$ is true) is covered $k + n$ times by $\{\{\varphi_1, \ldots, \varphi_n\}\}$, while $\bigvee_{J \subseteq \{1,\ldots,m\},\, |J|=k} \bigwedge_{j \in J} \varphi_j$ says that the whole space is covered $k$ times by $\{\{\varphi_1, \ldots, \varphi_n\}\}$; roughly speaking, $\{\{[\![\varphi_1]\!], \ldots, [\![\varphi_n]\!]\}\}$ is an $(n, k)$-cover of $([\![\varphi]\!], [\![\mathit{true}]\!])$. The conclusion of **L4** thus corresponds to the conclusion of **UP3**.

Instances of **Taut** include all formulas of the form $f \vee \neg f$, where $f$ is a likelihood formula. We could replace **Taut** by a simple collection of axioms that characterize propositional reasoning (see, for example, [Mendelson 1964]), but we have chosen to focus on aspects of reasoning about upper probability.

As in [Fagin, Halpern, and Megiddo 1990], the axiom **Ineq** includes "all valid formulas about linear inequalities."

Roughly speaking, an inequality formula is a formula of the form $a_1 x_1 + \cdots + a_n x_n \geq c$, over variables $x_1, \ldots, x_n$. The formula is said to be true if we satisfy the resulting inequality when we assign a real number to each variable of the formula. As usual, a formula is valid if it is true under every possible assignment of real numbers to variables. To get an instance of **Ineq**, we replace each variable $x_i$ that occurs in a valid formula about linear inequalities by a primitive likelihood term of the form $l(\varphi_i)$ (naturally each occurence of the variable $x_i$ must be replaced by the same primitive likelihood term $l(\varphi_i)$). As with **Taut**, we can replace **Ineq** by a sound and complete axiomatization for boolean combinations of linear inequalities. One such axiomatization is given in [Fagin, Halpern, and Megiddo 1990].

## 4 SOUNDNESS AND COMPLETENESS

A likelihood formula $f$ is *provable from* $F$ for $F$ a set of formulas if it can be proven using the axioms and rules of inferences, along with the formulas in $F$. In the special case where $F$ is empty, we say that $f$ is simply *provable*. An axiom system is *sound* if every provable formula is valid. An axiom system is *complete* if every valid formula is provable.

Our goal is to prove that $\mathbf{AX}^{up}$ is a sound and complete axiomatization for reasoning about upper probability. The soundness of $\mathbf{AX}^{up}$ is immediate from our earlier disscussion. Completeness is, as usual, harder. Unfortunately, the standard technique for proving completeness in modal logic, which involves considering maximal consistent sets and canonical structures (see, for example, [Popkorn 1994]) does not work. We briefly review the approach, just to point out the difficulties.

The standard approach uses the following definitions. A formula $\sigma$ is *consistent* with an axiom system **AX** if $\neg \sigma$ is not provable from **AX**. A finite set of formulas $\{\sigma_1, \ldots, \sigma_n\}$ is consistent with **AX** if the formula $\sigma_1 \wedge \cdots \wedge \sigma_n$ is consistent with **AX**; an infinite set of formulas is consistent with **AX** if all its finite subsets are consistent with **AX**. A *maximal* **AX**-consistent set of formulas $F$ is a set of formulas consistent with **AX** with the property that for any formula $\sigma \notin F$, $F \cup \{\sigma\}$ is not consistent with **AX**. Using just axioms of propositional logic, it is not hard to show that a **AX**-consistent set of formulas can be extended to a maximal **AX**-consistent set of formulas. To show that **AX** is a complete axiomatization with respect to some class of structures $\mathcal{M}$, we must show that every formula that is valid in every structure in $\mathcal{M}$ is provable in **AX**. To do this, it is sufficient to show that every **AX**-consistent formula $\sigma$ is satisfiable in $\mathcal{M}$. Typically, this is done by constructing what is called a *canonical structure* $M^c$ in $\mathcal{M}$ whose states are the maximal **AX**-consistent sets, and then showing that a formula $\sigma$ is satisfied in a world $w$ in $M^c$ iff $\sigma$ is one of the formulas in the canonical set associated with world $w$.

Unfortunately, this approach cannot be used to prove completeness here. To see this, consider the set of formulas:

$$F' = \{l(\varphi) \leq \frac{1}{n}, \ n = 1, 2, \ldots \} \cup \{l(\varphi) > 0\}.$$

This set is clearly $\mathbf{AX}^{up}$–consistent according to our definition, since every finite subset is satisfiable and $\mathbf{AX}^{up}$ is sound. It thus can be extended to a maximal $\mathbf{AX}^{up}$–consistent set $F$. However, the set $F'$ of formulas is not satisfiable: it is not possible to assign $l(\varphi)$ a value that will satisfy all the formulas at the same time. Hence, $F$ is not satisfiable. Thus, the canonical model approach, at least applied naively, simply will not work.

We take a different approach here, similar to the one taken in [Fagin, Halpern, and Megiddo 1990]. Specifically, we show that if a formula $f$ is $\mathbf{AX}^{up}$-consistent, then it is satisfiable in an upper probability structure. By a simple argument, we can easily reduce the problem to the case where $f$ is a conjunction of basic likelihood formulas and negations of basic likelihood formulas. Let $p_1, \ldots, p_N$ be the primitive propositions that appear in $f$. Observe that there are $2^{2^N}$ inequivalent propositional formulas over $p_1, \ldots, p_N$. The argument goes as follow. Let an *atom* over $p_1, \ldots, p_N$ be a formula of the form $q_1 \wedge \ldots \wedge q_N$, where $q_i$ is either $p_i$ or $\neg p_i$. There are clearly $2^N$ atoms over $p_1, \ldots, p_N$. Moreover, it is easy to see that any formula over $p_1, \ldots, p_N$ can be written in a unique way as a disjunction of atoms. There are $2^{2^N}$ such disjunctions, so the claim follows.

Let $\rho_1, \ldots, \rho_{2^{2^N}}$ be some canonical listing of the inequivalent formulas over $p_1, \ldots, p_N$. Without loss of generality, we assume that $\rho_1$ is equivalent to *true*, and $\rho_{2^{2^N}}$ is equivalent to *false*. Since every propositional formula over $p_1, \ldots, p_N$ is provably equivalent to some $\rho$, it follows that $f$ is provably equivalent to a formula $f'$ where each conjunct of $f'$ is of the form $\theta_1 l(\rho_1) + \cdots + \theta_{2^{2^N}} l(\rho_{2^{2^N}}) \geq \beta$. Note that the negation of such a formula has the form $\theta_1 l(\rho_1) + \cdots + \theta_{2^{2^N}} l(\rho_{2^{2^N}}) < \beta$ or, equivalently, $(-\theta_1) l(\rho_1) + \cdots + (-\theta_{2^{2^N}}) l(\rho_{2^{2^N}}) > -\beta$. Thus, the formula $f$ gives rise in a natural way to a system of inequalities of the form:

$$\begin{array}{rcl}
\theta_{1,1} l(\rho_1) + \cdots + \theta_{1,2^{2^N}} l(\rho_{2^{2^N}}) & \geq & \alpha_1 \\
\vdots & & \vdots \\
\theta_{r,1} l(\rho_1) + \cdots + \theta_{r,2^{2^N}} l(\rho_{2^{2^N}}) & \geq & \alpha_r \\
\theta'_{1,1} l(\rho_1) + \cdots + \theta'_{1,2^{2^N}} l(\rho_{2^{2^N}}) & > & \beta_1 \\
\vdots & & \vdots \\
\theta'_{s,1} l(\rho_1) + \cdots + \theta'_{s,2^{2^N}} l(\rho_{2^{2^N}}) & > & \beta_s.
\end{array} \quad (1)$$

We can express (1) as a conjunction of inequality formulas, by replacing each occurrence of $l(\rho_i)$ in (1) by $x_i$. Call this inequality formula $\overline{f}$.

If $f$ is satisfiable in some upper probability structure $M$, then we can take $x_i$ to be the upper probability of $\rho_i$ in $M$; this gives a solution of $\hat{f}$. However, $\hat{f}$ may have a solution without $f$ being satisfiable. For example, if $f$ is the formula $l(p) = 1/2 \wedge l(\neg p) = 0$, then $\hat{f}$ has an obvious solution; $f$, however, is not satisfiable in an upper probability structure, because the upper probability of the set corresponding to $p$ and the upper probability of the set corresponding to $\neg p$ must sum to at least 1 in all upper probability structures. Thus, we must add further constraints to the solution to force it to act like an upper probability.

**UP1–UP3** or, equivalently, the axioms **L1–L4**, describe exactly what additional constraints are needed. The constraint corresponding to **L1** (or **UP1**) is just $x_1 = 0$, since we have assumed $\rho_1$ is the formula *false*. Similarly, the constraint corresponding to **L2** is $x_{2^{2^N}} = 1$. The constraint corresponding to **L3** is $x_i \geq 0$, for $i = 1, \ldots, 2^{2^N}$. What about **L4**? This seems to require an infinite collection of constraints, just as **UP3** does.[5]

This is where **UPF3** comes into play. It turns out that, if $f$ is satisfiable at all, it is satisfiable in a structure with at most $2^N$ worlds, one for each atom over $p_1, \ldots, p_N$. Thus, we need to add only instances of **L4** where $k, m, n < B_{2^N}$ and $\varphi_1, \ldots, \varphi_m, \varphi$ are all among $\rho_1, \ldots, \rho_{2^{2^N}}$. Although this is a large number of formulas (in fact, we do not know exactly how large, since it depends on $B_{2^N}$, which we have not computed), it suffices for our purposes that it is a finite number. For each of these instances of **L4**, there is an inequality of the form $a_1 x_1 + \cdots + a_{2^{2^N}} x_{2^{2^N}} \geq k$. Let $\hat{f}$, the *inequality formula corresponding to $f$*, be the conjunction consisting of $\hat{f}$, together with all the inequalities corresponding to the relevant instances of **L4**, and the equations and inequalitites $x_1 = 0$, $x_{2^{2^N}} = 1$, and $x_i \geq 0$ for $i = 1, \ldots, 2^{2^N}$, corresponding to axioms **L1–L3**.

**Proposition 4.1:** *The formula $f$ is satisfiable in an upper probability structure iff the inequality formula $\hat{f}$ has a solution. Moreover, if $f$ has a solution, then $f$ is satisfiable in an upper probability structure with at most $2^{|f|}$ worlds.*

**Theorem 4.2:** *The axiom system $AX^{up}$ is sound and complete for upper probability structures.*

**Proof:** For soundness, it is easy to see that every axiom is valid for upper probability structures, including **L4**, which represents **UP3**.

For completeness, we proceed as in the discussion above. Assume that formula $f$ is not satisfiable in an upper probability structure; we must show that $f$ is $AX^{up}$–inconsistent. We first reduce $f$ to a canonical form. Let $g_1 \vee \cdots \vee g_r$ be a disjunctive normal form expression for $f$ (where each $g_i$ is a conjunction of basic likelihood formulas and their negations). Using propositional reasoning, we can show that $f$ is provably equivalent to this disjunction. Since $f$ is unsatisfiable, each $g_i$ must also be unsatisfiable. Thus, it is sufficient to show that any unsatisfiable conjunction of basic likelihood formulas and their negations is inconsistent. Assume that $f$ is such a conjunction. Using propositional reasoning and axiom **L5**, $f$ is equivalent to a likelihood formula $f'$ that refers to formulas $\rho_1, \ldots, \rho_{2^{2^N}}$. Since $f$ is unsatisfiable, so is $f'$. By Proposition 4.1, the the inequality formula $\hat{f}'$ corresponding to $f'$ has no solution. Thus, by **Ineq**, the formula $\neg f''$ that results by replacing each instance of $x_i$ in $\hat{f}'$ by $l(\rho_i)$ is $AX^{up}$–provable. All the conjuncts of $f''$ that are instances of axioms **L1–L4** are $AX^{up}$–provable. It follows that $\neg f'$ is $AX^{up}$–provable, and hence so is $\neg f$. ∎

## 5 DECISION PROCEDURE

Having settled the issue of the soundness and completeness of the axiom system $AX^{up}$, we turn to the problem of the complexity of deciding satisfiability. Recall the problem of satisfiability: given a likelihood formula $f$, we want to determine if there exists an upper probability structure $M$ such that $M \models f$. As we now show, the satisfiability problem is NP-complete, and thus no harder than satisfiability for propositional logic.

For the decision problem to make sense, we need to restrict our language slightly. If we allow real numbers as coefficients in likelihood formulas, we have to carefully discuss the issue of representation of such numbers. To avoid these complications, we restrict our language to allow only integer coefficients. Note that we can still express rational coefficients by the standard trick of "clearing the denominator". For example, we can express $\frac{2}{3} l(\varphi) \geq 1$ by $2l(\varphi) \geq 3$ and $l(\varphi) \geq \frac{2}{3}$ by $3l(\varphi) \geq 2$. Recall that we defined $|f|$ to be the length of $f$, that is, the number of symbols required to write $f$, where each coefficient is counted as one symbol. Define $||f||$ to be the length of the longest coefficient appearing in $f$, when written in binary. The size of a rational number $\frac{a}{b}$, denoted $||\frac{a}{b}||$, where $a$ and $b$ are relatively prime, is defined to be $||a|| + ||b||$.

A preliminary result required for the analysis of the decision procedure shows that if a formula is satisfied in some upper probability structure, it is satisfied in a structure $(\Omega, \Sigma, \mathcal{P}, \pi)$, which is "small" in terms of the number of states in $\Omega$, the cardinality of the set $\mathcal{P}$ of probability measures, and the size of the coefficients in $f$.

**Theorem 5.1:** *Suppose $f$ is a likelihood formula that is satisfied in some upper probability structure. Then $f$ is satisfied in a structure $(\Omega, \Sigma, \mathcal{P}, \pi)$, where $|\Omega| \leq |f|^2$, $\Sigma = 2^\Omega$ (every subset set of $\Omega$ is measurable), $|\mathcal{P}| \leq |f|$, $\mu(w)$* is a rational number such that $||\mu(w)||$ is $O(|f|^2||f|| + |f|^2 \log(|f|))$ for every world $w \in \Omega$ and $\mu \in \mathcal{P}$, and $\pi(w)(p) = \textbf{false}$ for every world $w \in \Omega$ and every primitive proposition $p$ not appearing in $f$.

---
[5]Although we are dealing with only finitely many formulas here, $\rho_1, \ldots, \rho_{2^{2^N}}$, recall that the formulas $\varphi_1, \ldots, \varphi_m$ in **L4** need not be distinct, so there are potentially infinitely many instances of **L4** to deal with.

**Theorem 5.2:** *The problem of deciding whether a likelihood formula is satisfiable in an upper probability structure is NP-complete.*

**Proof:** For the lower bound, it is clear that a given propositional formula $\varphi$ is satisfiable iff the likelihood formula $l(\varphi) > 0$ is satisfiable, therefore the satisfiability problem is NP-hard. For the upper bound, given a likelihood formula $f$, we guess a "small" satisfying structure $M = (\Omega, \Sigma, \mathcal{P}, \pi)$ for $f$ of the form guaranteed to exist by Theorem 5.1. We verify that $M \models f$ as follows. Let $l(\psi)$ be an arbitrary likelihood term in $f$. We compute $[\![\psi]\!]_M$ by checking the truth assignment of each $s \in \Omega$ and seeing whether this truth assignment makes $\psi$ true. We then replace each occurence of $l(\psi)$ in $f$ by $\max_{\mu \in \mathcal{P}} \{\sum_{s \in S_\psi} \mu(s)\}$ and verify that the resulting expression is true. ∎

## 6 CONCLUSION

We have considered a logic with the same syntax as the logic for reasoning about probability, inner measures, and belief presented in [Fagin, Halpern, and Megiddo 1990], with uncertainty interpreted as the upper probability of a set of probability measures. Under this interpretation, we have provided a sound and complete axiomatization for the logic. We further showed that the satisfiability problem is NP-complete (as it is for reasoning about probability, inner measures, and beliefs [Fagin, Halpern, and Megiddo 1990]), despite having to deal with probability structures with possibility infinitely many states and infinite sets of probability measures. The key step in the axiomatization involves finding a characterization of upper probability measures that can be captured in the logic. The key step in the complexity result involves showing that if a formula was satisfiable at all, it is satisfiable in a "small" structure, where the size of the state space as well as the size of the set of probability measures and the size of all probabilities involved, is polynomial in the length of the formula.

Given the similarity in spirit of the results for the various interpretations of the uncertainty operator (as a probability, inner measure, belief function, and upper probability), we conjecture that there is some underlying result from which all these results should follow. It would be interesting to make that precise.

An interesting generalization of the class of logics we have been discussing, namely logics with an uncertainty operator interpreted variously as a probability, an inner measure, a belief function, or an upper probability, is to consider expectations (or gambles). One can design a logic with an expectation operator instead of a likelihood operator, and again interpret the operator variously, as the expectation of a probability, an inner measure, a belief function, or an upper probability. (The work of Wilson and Moral [1994] is along those lines.) One advantage of working with expectation functions is that they are typically easier to characterize than the corresponding measures; for instance, the characterization of upper probabilities of gambles is much simpler than that of upper probabilities of events [Walley 1991]. Moreover, expectation functions lead in general to more expressive logics. We plan to report on this line of investigation in future work.

## Acknowledgments

Thanks to Dexter Kozen, Jon Kleinberg, and Hubie Chen for discussions concerning set covers. Vicky Weissman read a draft of this paper and provided numerous helpful comments. We also thank the anonymous UAI reviewers for their useful comments and suggestions. This work was supported in part by NSF grants IRI-96-25901 and IIS-0090145 and ONR grant N000140010341.